\documentclass[11pt,a4paper]{article}

\usepackage{xr}

\usepackage[hyperref]{emnlp-ijcnlp-2019}
\usepackage{times}
\usepackage{latexsym}
\usepackage{url}

\usepackage[shortlabels]{enumitem}

\aclfinalcopy %

\usepackage{tikz}
\usepackage{pgfplots}
\usepgfplotslibrary{fillbetween}
\usepgfplotslibrary{groupplots}
\usepackage{amsmath}
\usepackage{soul}
\usepackage{bbm}
\usepackage{setspace}
\usepackage{multirow}
\usepackage{bigdelim}
\usepackage{amsfonts}
\usepackage{balance}
\usepackage[T1]{fontenc}
\usepackage{amsmath,amsfonts,amssymb,amsthm}
\usepackage{dsfont}
\usepackage{xspace}

\usepackage{titlesec}
\titlespacing*{\section} {0pt}{1.2ex plus 1ex minus .2ex}{0.5ex plus .2ex}
\titlespacing*{\subsubsection}{0pt}{0.6ex plus 0.5ex minus .2ex}{0.3ex plus .2ex}
\titlespacing{\paragraph} {0pt}{0ex plus 0.1ex minus .2ex}{1em}
\makeatletter
\g@addto@macro\small{%
  \setlength\abovedisplayskip{-5pt}
  \setlength\abovedisplayshortskip{-5pt}
  \setlength\belowdisplayshortskip{-7pt}
  \setlength\belowdisplayskip{-7pt}
}
\makeatother

\usepackage{arydshln}
\newcommand{\dline}{\hdashline[0.5pt/1pt]}

\usepackage{algorithmicx} 
\usepackage[noend]{algpseudocode}
\usepackage{algorithm}
\usepackage[T1]{fontenc}

\algnewcommand\algorithmicdefinitions{\textbf{Definitions:}}
\algnewcommand\Definitions{\item[\algorithmicdefinitions]}
\renewcommand{\algorithmiccomment}[1]{{\color{gray}\raisebox{1px}{\texttt{\guillemotright}} #1}}
\algnewcommand{\LineComment}[1]{\Statex \hskip\ALG@thistlm \algorithmiccomment{#1}}
\algrenewcommand\alglinenumber[1]{\footnotesize #1:}
\algrenewcommand\algorithmicindent{1.0em}%
\makeatletter
\newcommand{\StatexIndent}[1][3]{%
  \setlength\@tempdima{\algorithmicindent}%
  \Statex\hskip\dimexpr#1\@tempdima\relax}

\newcommand{\nlstring}[1]{{\em #1}}

\newcommand{\naturals}{{\rm I\!N}}
\newcommand{\stdev}[2]{{#1}\begin{tiny}$\pm${#2}\end{tiny}}

\DeclareMathOperator{\vect}{vec}

\newcommand{\system}[1]{\textsc{#1}}

\newcommand{\thegame}{\textsc{CerealBar}\xspace}
\newcommand{\follower}{\mathrm{Follower}}
\newcommand{\leader}{\mathrm{Leader}}

\newcommand{\sentence}{\bar{x}}
\newcommand{\sentences}{\mathcal{X}}

\newcommand{\state}{s}
\newcommand{\states}{\mathcal{S}}

\newcommand{\istate}{\gamma}
\newcommand{\istates}{\Gamma}
\newcommand{\queue}{\bar{Q}}
\newcommand{\turntaker}{\alpha}
\newcommand{\stepsremaining}{\psi}
\newcommand{\maxsteps}{\Psi}

\newcommand{\action}{a}
\newcommand{\actions}{\mathcal{A}}
\newcommand{\worldactions}{\actions_w}
\newcommand{\act}[1]{\mathtt{#1}}
\newcommand{\doneaction}{\act{DONE}}

\newcommand{\transfunc}{\mathcal{T}}

\newcommand{\interaction}{\bar{I}}

\newcommand{\reward}{R}

\newcommand{\policy}{\pi}

\newcommand{\lingunet}{\system{LingUNet}\xspace}
\newcommand{\lingunetdepth}{L}

\newcommand{\featmap}{\mathbf{F}}
\newcommand{\embed}{\phi}
\newcommand{\kernel}{\mathbf{K}}

\newcommand{\sentencerep}{\mathbf{\sentence}}

\newcommand{\envembedding}{\mathbf{S}}
\newcommand{\envnumproperties}{P}
\newcommand{\position}{\rho}

\newcommand{\disttensor}{\mathbf{P}}
\newcommand{\rnn}{\textrm{RNN}}

\newcommand{\param}{\theta}

\title{Executing Instructions in Situated Collaborative Interactions}

\author{Alane Suhr \\
  Cornell University \\
  {\tt\footnotesize suhr@cs.cornell.edu} \\\And
  Claudia Yan \\
  IBM \\
  {\tt\footnotesize claudiab.yan@gmail.com} \\\And
  Charlotte Schluger\thanks{\hspace{5pt},$^{**}$: Equal contribution. All work done at Cornell.} \\
  Cornell University \\
  {\tt\footnotesize jes543@cornell.edu} \\\AND
  Stanley Yu$^*$ \\
  Columbia University \\
  {\tt\footnotesize stanley.yu@columbia.edu} \\\And
  Hadi Khader$^{**}$ \\
  Intel \\
  {\tt\footnotesize hadi.kh.khader@gmail.com} \\\AND
  Marwa Mouallem$^{**}$ \\
  IBM \\
  {\tt\footnotesize marwamouallem@gmail.com} \\\And
  Iris Zhang \\
  Facebook \\
  {\tt\footnotesize irisz@fb.com} \\\And
  Yoav Artzi \\
  Cornell University \\
  {\tt\footnotesize yoav@cs.cornell.edu}}

\date{}

\begin{document}

\maketitle

\begin{abstract}	

We study a collaborative scenario where a user not only instructs a system to complete tasks, but also acts alongside it. 
This allows the user to adapt to the system abilities by changing their language or deciding to simply accomplish some tasks themselves, and requires the system to effectively recover from errors as the user strategically assigns it new goals. 
We build a game environment to study this scenario, and learn to map user instructions to system actions. 
We introduce a learning approach focused on recovery from cascading errors between instructions, and  modeling methods to explicitly reason about instructions with multiple goals. 
We evaluate with a new evaluation protocol using recorded interactions and online games with human users, and observe how users adapt to the system abilities.

\end{abstract}

\section{Introduction}
\label{sec:intro}

\begin{figure}[t]
\footnotesize
\centering
\includegraphics[trim={115 187 196 136}, clip, width=0.98\linewidth]{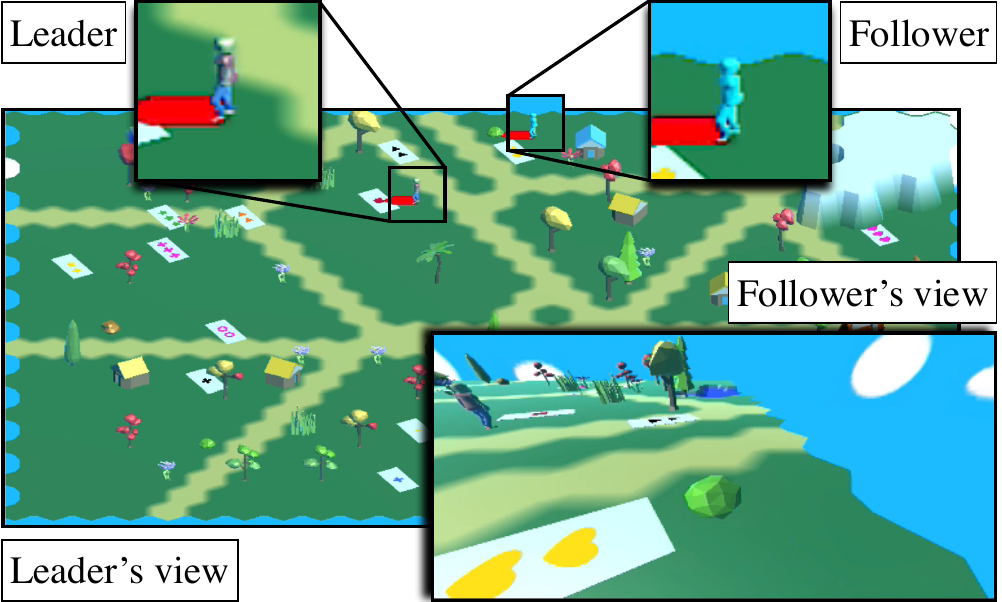}
\fbox{\begin{minipage}[]{0.95\linewidth}
\dots \\
$\sentence_3$: \nlstring{turn left and head toward the yellow hearts, but don't pick them up yet. I'll get the next card first.} \\
$\sentence_4$: \textbf{\nlstring{Okay, pick up yellow hearts and run past me toward the bush sticking out, on the opposite side is 3 green stars}}\\
\texttt{[Set made. New score: 4]} \\
\dots
\end{minipage}}
\caption{A snapshot from an interaction in \thegame. The current instruction is in bold. The large image shows the entire environment. This overhead view is available only to the leader. The follower sees a first-person view only (bottom right). The zoom boxes (top) show the leader and follower. 
}
\label{fig:intro}
\end{figure}

Sequential instruction scenarios commonly assume only the system performs actions, and therefore only its behavior influences the world state. 
This ignores the collaborative potential of such interactive scenarios and the challenges it introduces. 
When the user acts in the world as well, they can adapt to the system abilities not only by adopting simpler language, but also by deciding to accomplish tasks themselves. 
The system must then recover from errors as new instructions arrive and be robust to changes in the environment that are not a result of its own actions.

In this paper, we introduce \thegame, a collaborative game with natural language instruction, and design modeling, learning, and evaluation methods for the problem of sequential instruction following in collaborative interactions. 
In \thegame, two agents, a leader and a follower, move in a 3D environment and collect valid sets of cards  to earn points. 
A valid set is a set of three cards with distinct color, shape, and count. 
The game is turn-based, and only one player can act in each turn. 
In addition to collecting cards, the leader sends natural language instructions to the follower. 
The follower's role is to execute these instructions. 
Figure~\ref{fig:intro} shows a snapshot from the game where the leader plans to pick up a nearby card (red square) and delegates to the follower two cards, one close and the other much further away. Before that, the leader planned ahead and asked the follower to move in preparation for the next set. 
The agents have different skills to incentivize collaboration. 
The follower has more moves per turn, but can only see from first-person view, while the leader observes the entire environment but has fewer moves. 
This makes  natural language interaction key to success. 
We address the problem of mapping the leader instructions to follower actions. 
In addition to the collaborative challenges, this requires grounding natural language to resolve spatial relations and references to objects, reason about dependencies on the interaction history, react to the changing environment as cards appear and disappear, and generate actions. 

\thegame requires reasoning about the changing environment (e.g., when selecting cards) and instructions with multiple goals (e.g., selecting multiple cards). 
We build on the Visitation Prediction Network model~\cite[VPN;][]{Blukis:18visit-predict}, which casts planning as mapping instructions to the probability of visiting positions in the environment. 
Our new model generalizes the planning space of VPN to reason about intermediate goals and obstacles, and includes recurrent action generation for trajectories with multiple goals.

We collect 1,202 human-to-human games for training and evaluation. 
While our model could be trained from these recorded games only, it would often fail when an instruction would start at the wrong position because of an error in following the previous one. 
We design a learning algorithm that dynamically augments the data with examples that require recovering from such errors, and train our model to distinguish such recovery reasoning from regular instruction execution.

Evaluation with recorded games poses additional challenges. 
As agent errors  lead to unexpected states, later instructions become invalid. 
Because measuring task completion from such states is meaningless,  we propose \emph{cascaded evaluation}, a new evaluation protocol that starts the agent at different points in the interaction and measures how much of the remaining instructions it can complete. 
In contrast to executing complete sequences or single instructions, this method allows to evaluate all instructions while still measuring  the effects of error propagation.

We evaluate using  both static recorded games and live interaction with human players. 
Our human evaluation shows users adapt to the system and use the agent effectively, scoring on average 6.2 points, compared to 12.7 for human players. 
Our data, code, and demo videos are available at \href{http://lil.nlp.cornell.edu/cerealbar/}{lil.nlp.cornell.edu/cerealbar/}.

\section{Setup and Technical Overview}
\label{sec:overview}

We consider a setup where two agents, a leader and a follower, collaborate. Both execute actions in a shared environment. The leader, additionally, instructs the follower using natural language. 
The leader goal is to maximize the task reward, and the follower goal  is to execute leader instructions. 
We consider a turn-based version, where at each turn only one agent acts. %
We instantiate this scenario in \thegame, a navigation card game (Figure~\ref{fig:intro}), where a leader and follower move in an environment selecting cards to complete sets.\footnote{The name \thegame does not carry special meaning. It was given to the project early on, and we came to like it. Our game is inspired by \href{https://en.wikipedia.org/wiki/Set_(card_game)}{the card game Set}.}

\paragraph{\thegame Overview}

The objective of \thegame is to earn points by selecting valid sets of cards. 
A valid set has three cards with distinct color, shape, and count.
When the only cards selected in the world form a valid set, the players receive a point, the selected cards disappear, three new cards are added randomly, and  the number of remaining turns increases. 
The increase in turns decays for each set completed. 
An agent stepping on a card flips its selection status. 
The players form sets together. 
The follower has more steps per turn than the leader. 
This makes using the follower critical for success. 
The follower only sees  a first-person view of the environment, preventing them from planning themselves, and requiring instructions to be sensible from the follower's perspective. 
The leader chooses the next target set, plans which of the two players should get which card, and instructs the follower.
The follower  can not respond to the leader, and should not plan themselves, or risk sabotaging the leader's plan, wasting moves and lowering their potential score. 
Followers mark an instruction as finished before observing the next one. This provides alignment between instructions and follower actions.
In contrast to the original setup that we use for data collection, in our model (Section~\ref{sec:model}), we assume the follower has full observability, leaving the challenge of partial observability for future work. 
Appendix~\ref{sec:sup:game} provides further game design details.

\paragraph{Problem Setup}

We distinguish between the world state  and the interaction state. 
Let $\states$ be the set of all world states, $\istates$ be the set of all interaction states, and $\sentences$ be the set of all natural language instructions. 
A world state $\state \in \states$ describes the current environment. 
In \thegame, the world state describes the spatial environment, the location of cards, whether they are selected or not, and the location of the agents. 
An interaction state $\istate \in \istates$ is a tuple $\langle \queue, \turntaker, \stepsremaining \rangle$. 
The first-in-first-out queue $\queue = [\sentence_q,\dots,\sentence_{q'}]$ contains the  instructions $\sentence_i \in \sentences$ available to execute. The current  instruction  is the left-most instruction $\sentence_q$. 
The current turn-taker $\turntaker \in \{\leader, \follower\}$ indicates the agent currently executing actions, and  $\stepsremaining \in \naturals_{\geq 0}$ is the number of steps remaining in the current turn.

At each time step, the current turn-taker agent takes an action. 
An action may be the leader issuing an instruction, or either agent performing an action in the environment. 
Let $\actions = \worldactions \cup \{ \doneaction \}\cup \sentences$ be the set of all actions.
The set $\worldactions$ includes the actions available to the agents in the environment. 
In \thegame, this includes moving forward or backward, and turning left or right. 
Moving onto a card flips it selection status.
$\doneaction$ indicates completing the current instruction for the follower or ending the turn for the leader. 
An instruction action $\action = \sentence \in \sentences$ can only be taken by the leader and adds the instruction $\sentence$  to the queue $\queue$. 
The effect of each action is determined by the transition function $\transfunc :\states \times \istates \times \actions \rightarrow \states \times \istates$, which is formally defined in Appendix~\ref{sec:sup:transfunc}. 
Only world actions $\action \in \worldactions$ decrease the remaining steps $\stepsremaining$.

The goal of the leader is to maximize the total reward of the interaction. 
An interaction $\interaction = \langle (\state_1, \istate_1, \action_1), \dots, (\state_{|\interaction|}, \istate_{|\interaction|}, \action_{|\interaction|}) \rangle$ is a sequence of state-action tuples, where $\transfunc\left(\state_{i}, \istate_{i}, \action_{i} \right)=(\state_{i+1}, \istate_{i+1})$.
The reward function $\reward : \states \times \actions \rightarrow \mathbb{R}$ assigns a numerical reward to a world state and an action. 
The total reward of an interaction $\interaction$ is $\sum_{i=0}^{|\interaction|}\reward(\state_i, \action_i)$. 
In \thegame, the agents receive a reward of 1 when a valid set  is selected. 

\paragraph{Task}

Our goal is to learn a follower policy to execute the leader instructions. 
At time $t$, given the current world and interaction states $\state_t$ and $\istate_t$, and the interaction so far $\interaction_{<t}$, the follower policy $\policy(\state_t, \istate_t, \interaction_{<t})$ predicts the next action  $\action_t$.

\paragraph{Model}

We decompose the follower policy  $\policy(\state_t, \istate_t, \interaction_{<t})$ to predicting a set of distributions over positions in the environment, including positions to visit, intermediate goals (e.g., cards to select), positions to avoid (e.g., cards not to touch), and positions that are not passable. These distribution are used in a second stage to generate a sequence of actions.  
Section~\ref{sec:model} describes the model.

\paragraph{Learning}
We assume access to a set of $N$ recorded interactions $\{ \interaction^{(i)} \}^N_{i=1}$, and create examples where each instruction is paired with a sequence of state-action tuples. 
We maximize the action-level cross entropy objective, and use two auxiliary objectives (Section~\ref{sec:learning}). 
We first train each stage of the model separately, and then fine-tune them jointly. During fine-turning, we  continuously generate additional examples using model failures. These examples help the agent to learn how to recover from errors in prior instructions.

\paragraph{Evaluation}

We measure  correct execution of instructions and the overall game reward. 
We assume access to a test set of $M$ recorded interactions $\{ \interaction^{(i)} \}^M_{i=1}$. 
We measure instruction-level and interaction-level performance, 
and develop \emph{cascaded evaluation}, an evaluation protocol that provides a more graded measure than treating each interaction as a single example, while still accounting for error propagation (Section~\ref{sec:cascaded}). 
Finally, we  conduct online evaluation with human leaders.

\section{Related Work}
\label{sec:related}

Goal-driven natural language interactions have been studied in various scenarios, including dialogue where only one side acts in the world~\cite{Anderson:91,Williams:13dstc,Vlachos:14,Devries:18,Kim:19codraw,Hu:19minirts}, coordination for agreed selection of an object~\cite{He:17dialogue, Udagawa:19}, and negotiation~\cite{Lewis:17negotiation,He:18negotiation}. 
We focus on collaborative interactions where both the user and the system perform sequences of actions  in the same environment. 
This allows the user to adapt to the language understanding ability of the system and balance between delegating goals to it and accomplishing them themselves. 
For example, a user may decide to complete a short but hard-to-describe task and delegate to the system a long but easy-to-describe one. 
In prior work, in contrast, recovery is limited to users paraphrasing their requests. 
The Cards corpus~\cite{Djalali11:cards-qud,Djalali11:cards-preference,Potts:12} was used for linguistic analysis of   collaborative  bi-directional language interaction. 
The structure of collaborative interactions was also studied using Wizard-of-Oz studies~\cite{Lochbaum:98,Sidner:00,Koulouri:09}. 
In contrast, we focus on building agents that  follow instructions. 
\citet{Ilinykh:19meetup} present a corpus for the related task of natural language coordination in navigation.  
Collaboration has also been studied for emergent communication~\cite[e.g.,][]{Andreas:17,Evtimova:17}.

Understanding sequences of natural language utterances has been addressed using semantic parsing~\cite[e.g.,][]{Miller:96,MacMahon:06,Chen:11,Artzi:13,Artzi:14,Long:16context,Iyyer:17seq-qa,Suhr:18context,Arkin:17,Broad:17}. 
Interactions were also used for semantic parser induction~\cite{Artzi:11,Thomason:15robotdialog,Wang:16games}. 
These methods require hand-crafted symbolic meaning representation, while we use low-level actions~\cite{Suhr:18situated}. 
The interactions in our environment interleave actions of both agents with leader utterances, an aspect not addressed by these methods.  
Executing single instructions has been widely studied~\cite[e.g.,][]{Tellex:11,Duvallet:13,Misra:17instructions,Misra:18goalprediction,Anderson:17,Blukis:18drone,Blukis:18visit-predict,Chen:19touchdown}. 
The distinction we make between  actions specified in the instruction and implicit recovery actions is similar to how \citet{Artzi:13} use implicit actions  for single instructions. 
Finally, our model is based on the VPN model of \citet{Blukis:18visit-predict}. While we assume full observability, their original work did not. This indicates that our model is likely to generalize well to partially observable scenarios.

\section{Model}
\label{sec:model}

\begin{figure*}
    \centering
    \includegraphics[width=\textwidth,trim={92pt 274pt 99pt 72pt},clip=true]{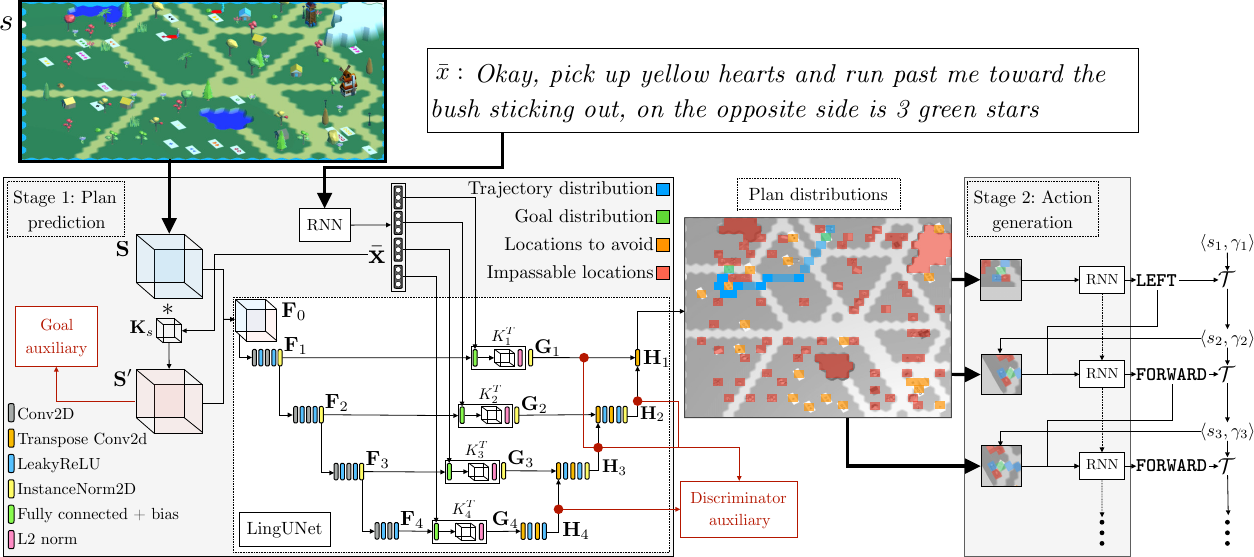}
    \caption{Illustration of the model architecture. Given the instruction $\sentence$ and the world state $\state$, we compute  $\featmap_0$ from the embeddings of the instruction $\sentencerep$ and environment  $\envembedding$. We use $\lingunet$ to predict four distributions, which are visualized over the map (grayscaled to emphasize the distributions). We show three action generation steps. Each step receives the map cropped around the agent and the previous action, and outputs the next action. }
    \label{fig:diagram}
\end{figure*}

We use a two-stage model for the follower policy $\policy(\state_t, \istate_t, \interaction_{<t})$, where $\state_t$ is a world state, $\istate_t$ is an interaction state, and $\interaction_{<t}$ is the interaction history. 
The instruction $\sentence$ that is the first in the queue $\queue_t$, which is part of $\istate_t$, is the currently executed instruction. 
In our model, we assume the follower observes the entire environment. 
First, we map $\sentence$ and $\state_t$ to distributions over locations in the environment, including what locations to visit and what are the goals. 
These distributions are considered as an execution plan, and are used to generate a sequence of actions in the second stage. 
The distribution can also be used to easily  easily visualize the agent plan. 
The first stage is used when starting a new instruction, and the predicted distributions are re-used for all actions for that instruction. 
Figure~\ref{fig:diagram} illustrates the architecture and the distributions visualization. 
The two-stage approach was introduced by \citet{Blukis:18visit-predict}. 
We generalize its planning space and add a recurrent action generator for execution.

\paragraph{Input Representation}

The inputs to the first stage are the instruction  $\sentence$ and the world state $\state_t$. 
We generate feature maps for both. 
We use a learned embedding function $\embed^{\sentences}$ and a bi-directional recurrent neural network~\cite[RNN;][]{Elman:90rnn} with a long short-term memory cell~\cite[LSTM;][]{Hochreiter:97lstm} $\rnn^\sentences$ to map $\sentence$ to a vector $\sentencerep$. 
The world state $\state_t$ is a 3D tensor that encodes the properties of each position. 
The dimensions of $\state_t$ are $\envnumproperties\times W\times H$, where $\envnumproperties$ is the number of properties, and $W$ and $H$ are the environment width and height. 
Each  of the $W\times H$ positions is represented in $\state_t$ as a binary vector of length $\envnumproperties$. 
For example, a position with a red hut will have 1's for the \emph{red} and \emph{hut} dimensions and 0's for all other dimensions.  
We map the world state to a tensor feature map $\featmap_0$ by embedding $\state_t$ and processing it using the text representation $\sentencerep$.
We use a learned embedding function $\embed^{\states}$ to map each position vector to a dense embedding of size $N_s$ by summing embeddings of each of the position's properties.
The embeddings are combined to a tensor $\envembedding$ of dimension $N_s \times W \times H$ representing a featurized global view of the environment. 
We create a text-conditioned state representation by creating a kernel $\kernel_s$
 and   convolving with it over $\envembedding$. 
We use a linear transformation to create $\kernel_s = \mathbf{W}_s \sentencerep + \mathbf{b}_s$, 
where $\mathbf{W}_s$ and $\mathbf{b}_s$ are learned weights. 
We reshape $\kernel_s$ to a $1\times 1$ convolution kernel with $N_{s'}$ output channels, and compute $\envembedding' = \envembedding * \mathbf{K}_s$. 
We concatenate $\envembedding$ and $\envembedding'$ along the channel dimension and rotate and center so the follower position is at center pixel to generate $\featmap_0$.\footnote{Appendix~\ref{app:sec:lingunet} describes the relationship between the environment representations and the agent's initial and current orientation.}

\paragraph{Stage 1: Plan Prediction}

We treat plan generation as predicting distributions over positions $\position$ in the environment. There are $W \times H$ possible positions. 
We predict four distributions: 
(a) \mbox{$p(\position \mid \state_t, \sentence)$}, the probability of visiting  $\position$  while executing the instruction $\sentence$; 
(b) \mbox{$p(\mathrm{GOAL} = 1 \mid \position, \state_t, \sentence)$}, the binary probability that  $\position$  is a goal (i.e., $\mathrm{GOAL} = 1$ when containing a card to select); 
(c) \mbox{$p(\mathrm{AVOID} = 1 \mid \position, \state_t, \sentence)$}, the binary probability that the agent must not pass in $\position$ (i.e., $\mathrm{AVOID} = 1$ when it contains a card that should not be touched); 
and (d) \mbox{$p(\mathrm{NOPASS} = 1\mid \position, \state_t, \sentence)$}, the binary probability the agent cannot pass in $\position$ (i.e., $\mathrm{NOPASS} = 1$ when it contains another object). 

We use $\lingunet$~\cite{Misra:18goalprediction} to predict the distributions. The inputs to $\lingunet$ are the instruction embedding $\sentencerep$ and featurized world state $\featmap_0$, which is relative to the agent's frame of reference. 
The output are four matrices, each of dimension $W \times H$ corresponding to the environment. 
$\lingunet$ is formally defined in \citet{Misra:18goalprediction} and  Appendix~\ref{app:sec:lingunet}.
Roughly speaking, $\lingunet$ reasons about the environment representation $\featmap_0$ at $\lingunetdepth$ levels.
First, $\featmap_0$ is used to generate feature maps of decreasing size $\featmap_j$, $j=1\dots\lingunetdepth$ using a series of convolutions. 
We create convolution kernels from the instruction representation $\sentencerep$, and apply them to the feature maps $\featmap_j$ to generate text-conditioned feature maps $\mathbf{G}_j$. 
Finally, feature maps of increasing size $\mathbf{H}_j$ are generated using a series of $\lingunetdepth$ deconvolutions. 
The last deconvolution generates a tensor of size $4 \times W \times H$ with a channel for each of the four distributions. 
We use a softmax over one channel to compute $p(\position \mid \state_t, \sentence)$. 
Because the other distributions are binary, we use a sigmoid on each value independently for the other channels. 
When computing $p(\mathrm{GOAL} = 1 \mid \position, \state_t, \sentence)$ and $p(\mathrm{AVOID} = 1 \mid \position, \state_t, \sentence)$ we mask positions without objects that can be changed (i.e., positions without cards) to assign them zero probability.

\paragraph{Stage 2: Action Generation}

We use the four distributions to generate a sequence of actions. 
We concatenate the distributions channel-wise to a tensor $\disttensor \in \mathbb{R}^{4 \times W \times H}$.
We use a forward LSTM RNN to predict a sequence of actions. 
At each prediction step $t$, we rotate, transform, and crop $\disttensor$ to generate the egocentric tensor $\disttensor'_t \in \mathbb{R}^{N' \times C \times C}$, where the agent is always at the center and facing in the same direction, such that $\disttensor'_t$ is relative to the agent's current frame of reference.
The input to the action generation RNN at time $t$ is: 

\begin{small}
\begin{eqnarray*}
  \mathbf{p}'_t &=& \vect(\textsc{Norm}(\textsc{ReLU}(\textsc{CNN}^P(\disttensor'_t)))) \\
  \mathbf{p}_t &=& \textsc{ReLU}(\mathbf{W}^P_2[\mathbf{p}'_t; \textsc{ReLU}(\mathbf{W}^P_1\mathbf{p}'_t + \mathbf{b}^P_1)] + \mathbf{b}^P_2)\;\;,
\end{eqnarray*}
\end{small}

\noindent
where $\textsc{CNN}^P$ is a convolutional layer, $\textsc{ReLU}$ is a non-linearity, $\textsc{Norm}$ is instance normalization~\cite{Ulyanov2017:instancenorm}, and $\mathbf{W}^P_1$, $\mathbf{W}^P_2$, $\mathbf{b}^P_1$, $\mathbf{b}^P_2$ are learned weights. 
The action probability is: 

\begin{small}
\begin{eqnarray}
\nonumber  \mathbf{h}_t &=& \rnn^\actions(\mathbf{h}_{t-1}, [\embed^{\actions}(\action_{t-1}); \mathbf{p}_t]) \\ 
\label{eq:prob_act}  p(\action) &\propto & \exp(\mathbf{W}^\actions[\mathbf{h}_t; \mathbf{p}_t] + \mathbf{b}^\actions)\;\;,
\end{eqnarray}
\end{small}

\noindent
where $\rnn^\actions$ is an LSTM RNN, $\embed^\actions$ is a learned action embedding function,  $\action_0$ is a special $\mathtt{START}$ action, and $\mathbf{W}^\actions$ and $\mathbf{b}^\actions$ are learned. 
During inference, we assign zero probabilities to actions $a$ when $\transfunc_w(s_t, a)$ is invalid (Appendix~\ref{sec:sup:transfunc}), for example when an agent would move into an obstacle.

\section{Learning}
\label{sec:learning}

We assume access to a set of $N$ recorded interactions $\{ \interaction^{(i)} \}^N_{i=1}$. 
We generate instruction-level examples $\mathcal{D}  = \bigcup^N_{i=1}\{ \interaction^{(i,j)}\}^{M^{(i)}}_{j=1}$, where $M^{(i)}$ is the number of examples from $\interaction^{(i)}$. 
Each $\interaction^{(i,j)} = \langle (\state^{(i,j)}_1, \istate^{(i,j)}_1, \action^{(i,j)}_1), \dots, (\state^{(i,j)}_{k}, \istate^{(i,j)}_{k}, \action^{(i,j)}_{k}) \rangle$   is a subsequence of tuples in $\interaction^{(i)}$, where $\action^{(i,j)}_1$ is the first action the follower takes after observing the $j$-th instruction in $\interaction^{(i)}$, and $\action^{(i,j)}_{k}$ is the $\doneaction$ action completing that instruction.
We first estimate the parameters for plan prediction $\param_1$ and action generation $\param_2$ separately (Section~\ref{sec:learn:pretrain}), and then fine-tune jointly with data augmentation (Section~\ref{sec:learn:finetune}).

\subsection{Pretraining}
\label{sec:learn:pretrain}

\paragraph{Stage 1: Plan Prediction}
The input of Stage 1 is the world state $\state_1$ and the instruction $\sentence$ at the head of the queue $\queue$.\footnote{We omit example indices for succinctness.} 
We generate labels for the four output distributions using  $\interaction^{(i,j)}$. 
The visitation distribution $p(\position \mid \state_1, \sentence)$ label is proportional to number of states $s_t \in \interaction^{(i,j)}$ where the follower is in position $\position$. 
The goal and avoidance distributions model how the agent plans to manipulate parts of its environment to achieve the specified goals, but avoid manipulating other parts.
In \thegame, this translates to changing the status of cards, or avoiding doing so. 
For $p(\mathrm{GOAL} = 1 \mid \position, \state_1, \sentence)$, we set the label to 1 for all $\position$ that contain a card that the follower changed its selection status in $\interaction^{(i,j)}$, and 0 for all other positions. 
Similarly, for the avoidance distribution $p(\mathrm{AVOID} = 1 \mid \position, \state_1, \sentence)$, the label is 1 for all $\position$ that have cards that the follower does not change during the interaction $\interaction^{(i,j)}$. 
Finally, for $p(\mathrm{NOPASS} = 1\mid \position, \state_1, \sentence)$, the label is 1 for all positions the agent cannot move onto, and zero otherwise.
We define four cross-entropy losses: visitation $\mathcal{L}_V$, goal $\mathcal{L}_G$, avoidance $\mathcal{L}_A$, and no passing $\mathcal{L}_P$.
We also use an auxiliary cross-entropy goal-prediction loss $\mathcal{L}_{G'}$ using a probability $p'_G(\mathrm{GOAL} = 1 \mid \position, \state_1, \sentence)$ we predict from the pre-$\lingunet$ representation $\envembedding'$ by classifying each position. The complete loss is a weighted sum with coefficients:\footnote{Additional details are in Appendix~\ref{sec:sup:learn}.}  

\begin{small}
\begin{align*}
    \mathcal{L}_1(\param_1) = &\lambda_V \mathcal{L}_V(\param_1) + \lambda_G \mathcal{L}_G(\param_1) +\lambda_A \mathcal{L}_A(\param_1) \\
    &+ \lambda_P\mathcal{L}_P(\param_1) + \lambda_{G'}\mathcal{L}_{G'}(\param_1)\;\;.
\end{align*}
\end{small}

\paragraph{Stage 2: Action Generation}

We use the gold distribution to create the input $\disttensor$, and optimize towards the annotated set of actions using teacher forcing~\cite{Williams:89teacherforcing}. 
We compute the loss only over actions taken by the follower: 

\begin{small}
\begin{equation*}
    \mathcal{L}_2(\param_2) = -\textstyle\sum_{t=1}^n \mathds{1}_{\alpha_t = \follower}\log p(\action_t)\;\;,
\end{equation*}
\end{small}

\noindent
where $p(\action_t)$ is computed by Equation~\ref{eq:prob_act}.

\subsection{Fine-tuning with Example Aggregation}
\label{sec:learn:finetune}

Simply combining the separately-trained networks together results in low performance.
We perform additional fine-tuning with the two stages combined, and  introduce  a data augmentation method to learn to recover from error propagation.

\paragraph{Error Propagation}

Executing a sequence of instructions is susceptible to error propagation, where an agent fails to correctly complete an instruction, and because of it also fails on the following ones. 
While the collaborative, turn-switching setup allows the leader to adjust their plan following a follower mistake, leaders often strategically issue multiple instructions to use the available follower steps optimally. 
Given an agent failure, subsequent instructions may not align with the state of the world resulting from the follower's error. 
In supervised learning, we do not have the opportunity to learn to recover from such errors, even when it is relatively simple. 
This usually requires exploration. 
However, conventional frameworks like reinforcement learning (RL) or imitation learning (IL) are poorly suitable. 
In a live interaction, when an agent makes a mistake (e.g., selecting the wrong card), the leader is likely to adjust their actions. 
Because of this, in a recorded interaction, which contains the leader actions following a correct execution, it is not possible to reliably compute an RL reward for states following erroneous executions. 
For similar reasons, we cannot compute an IL oracle.

We identify two classes of  erroneous states in \thegame: 
(a) not selecting the correct set of cards; 
and (b) finishing with the right card selection, but stopping at the wrong position.\footnote{See Appendix~\ref{sec:app:learning} for further discussion of the two cases.} 
Case (a) requires to modify the model, for example to know when to skip instructions that refer to a state that is no longer possible. We leave this case for future work.
We address case (b) by augmenting the data with new examples that are aggregated during learning. 
Our process is similar to \textsc{DAgger}~\cite{Ross:11dagger}. 
We alternate between: (a) collecting new training examples using a heuristic oracle, and (b) performing model updates.
We generate training examples that demonstrate  recovery by starting in an incorrect initial position for an instruction, having arrived there by executing the previous instruction. 
We train our model to distinguish between the reasoning required for generating implicit actions to correct errors, and explicit actions directly mentioned in the instruction.

\paragraph{Learning with Example Aggregation}

We alternate between aggregating a new set of recovery examples $\mathcal{D}'$ and updating our parameters. 
At each epoch, we first use the current policy to create new training examples. 
We run inference for each example $\interaction^{(i,j)}$ in $\mathcal{D}$, the original training set, using the current policy.\footnote{We do not perform inference for the last instruction in an interaction, as there is no subsequent example for which to generate a new example.}
We compare the state $\state'$ at the end of execution to the final state in $\interaction^{(i,j)}$ to generate an error-recovery example $\interaction'^{(i,j+1)}$ for the subsequent example $\interaction^{(i,j+1)}$. 
We only generate such examples if the position or rotation of the agent are different, and there are no other difference between the states. 
Starting from $\state'$, we generate the shortest-path sequence of actions that: (a) changes the cards as specified in $\interaction^{(i,j+1)}$, and (b) executes  $\doneaction$ in the same position as in $\interaction^{(i,j+1)}$.
We then create $\interaction'^{(i,j+1)}$ using $\interaction^{(i,j+1)}$ and the new sequence of state-action pairs, and add it to $\mathcal{D}'$.\footnote{Appendix~\ref{sec:app:learning} describes this process.}

Given the original set of examples $\mathcal{D}$ and the aggregate examples $\mathcal{D}'$ we update our model parameters. 
We randomly sample without replacement at most $\sum_{i=1}^N M^{(i)}$ examples, the size of $\mathcal{D}$, from $\mathcal{D}'$.
We use all the examples in $\mathcal{D}$ and the sampled examples to do a single parameter update epoch. 
We limit the number of examples from $\mathcal{D}'$ to maintain the effect of the original data.

\paragraph{Optimizing with Implicit Action Prediction}

The examples we generate during aggregation often include sequences of state-action pairs that do not align with the instruction, for example when a mentioned spatial relation is incorrect from the new starting position. 
Such examples require reasoning differently about the text and world state than with the original data. 
We identify such examples in $\mathcal{D}'$ by comparing their follower starting position to the starting position in the original corresponding example in $\mathcal{D}$. 
If the distance is over two, we treat the examples as requiring implicit actions~\cite{Artzi:13}. All other examples, including all original examples in $\mathcal{D}$ are considered as not requiring implicit reasoning. 
We encourage the model to reason differently about these examples with a discriminator that classifies if the example requires implicit reasoning or not using the internal activations of $\lingunet$.

The discriminator classifies each of the $\lingunetdepth$ layers in $\lingunet$ for implicit reasoning. 
The goal is to encourage implicit reasoning at all levels of reasoning in the first stage. 
The probability of implicit reasoning for each $\lingunet$ layer $l$ is:

\begin{small}
\begin{eqnarray*}
  p(\mathrm{IMPLICT} = 1 \mid l, \state_1, \sentence) =& \\ &\hspace{-2cm} \begin{cases}  \sigma(\textsc{AvgPool}(\mathbf{G}_1 * \kernel^{\rm IMP}_1)) & l=1 \\
     \sigma(\textsc{AvgPool}(\mathbf{H}_l * \kernel^{\rm IMP}_l))  & l>1\end{cases} \;\;,
\end{eqnarray*}
\end{small}

\noindent
where $\kernel^{\rm IMP}_l$ are $1 \times 1$ learned kernels and $\textsc{AvgPool}$ does average pooling. 
We define a cross-entropy loss  $\mathcal{L}_{\rm IMP}$ that averages across  the $\lingunetdepth$ layers. 
The complete fine-tuning loss is: 

\begin{small}
\begin{equation*}
    \mathcal{L}(\param_1, \param_2) = \mathcal{L}_1(\param_1) + \mathcal{L}_2(\param_2) + \lambda_{\rm IMP}\mathcal{L}_{\rm IMP}(\param_1)\;\;.
\end{equation*}
\end{small}

\section{Cascaded Evaluation}
\label{sec:cascaded}

Sequential instruction scenarios are commonly evaluated using  recorded interactions by executing individual instructions or executing complete interactions starting from their beginning~\cite[e.g.,][]{Chen:11,Long:16context}. 
Both have limitations.
Instruction-level metrics ignore error propagation, and do not accurately reflect the system's performance. 
In contrast, interaction-level metrics do consider error propagation and capture overall system performance well.
However, they poorly utilize the test data, especially when performance is relatively low. 
When early failures lead to unexpected world states, later instructions become impossible to follow, and measuring performance on them is meaningless. 
For example, with our best-performing model, 82\% of development instructions become impossible due cascading errors when executing complete interactions.

The two measures may also fail to distinguish models. 
For example, consider an interaction with three instructions. 
Two models, A and B, successfully execute the third instruction in isolation, but fail on the two others. They also both fail when executing the entire interaction starting from the beginning. 
According to common measures, the models are equal. 
However, if model B can actually recover from failing on the second instruction to successfully execute the third, it means it is better than model A. 
Both metrics fail to reflect this.

We propose \emph{cascaded evaluation}, an evaluation protocol for sequential instruction using static corpora. 
Our method utilizes all instructions during testing, while still accounting for the effect of error propagation. 
Unlike instruction-level evaluation, cascaded evaluation executes the instructions in sequence. 
However, instead of starting of starting only from the start state of the first instruction, we create separate examples for starting from the starting state of each instruction in the interaction and continuing until the end of the interaction. For example, given a sequence of three instructions $\langle1,2,3\rangle$ we will create three examples: $\langle 1, 2, 3\rangle$, $\langle 2,3\rangle$, and $\langle 3 \rangle$.
To evaluate performance in \thegame, we compute two statistics using cascaded evaluation: the proportion of the remaining instructions followed successfully, and the proportion of potential points scored. We only consider the remaining instructions and points left to achieve in the example. For example, for the sequence $\langle 2,3\rangle$, we will subtract any points achieved before the second instruction to compute the proportion of potential points scored. 
Appendix~\ref{sec:sup:cascaded} describes cascaded evaluation formally.

\section{Experimental Setup}

\paragraph{Data}
We collect 1,202 human-human interactions using Mechanical Turk, split into train (960 games), development (120), and test (122). 
Appendix~\ref{sec:sup:data} details  data collection  and statistics. 

\paragraph{Recorded Interactions Metrics}
We evaluate instruction-level, interaction-level, and cascaded (Section~\ref{sec:cascaded}) performance. 
We allow the follower ten steps per turn, and interleave the actions taken by the leader during each turn in the recorded interaction. Instruction execution often crosses turns. 
At the instruction-level, we evaluate the mean \emph{card state accuracy} comparing the state of the cards after inference with the correct card state, \emph{environment state accuracy} comparing both cards and the agent's final position,  and \emph{action sequence accuracy} comparing the generated action sequence with the correct action sequence.
For complete interactions, we measure mean \emph{full game points}. 
Finally, for cascaded evaluation, we measure the mean proportion  of instructions correctly executed and of possible points scored.

\paragraph{Human Evaluation}
We perform evaluation with human leaders, comparing our model and human followers. 
Workers are told they will work with a human or an automated follower, but are not told which in each game.
We evaluate both human (105 games) and automated agents at the same time (109 games).
We evaluate the game scores, and also elicit free-form feedback. 

\begin{table*}[th!]
    \centering\footnotesize
    \begin{tabular}{|l||c|c|c||c||c|c|} \hline
       \multicolumn{1}{|c||}{\multirow{2}{*}{\textbf{System}}}  &  \textbf{Card} & \textbf{Env.} & \textbf{Action Seq.} & \textbf{Full Game} & \textbf{Prop. Instr.} & \textbf{Prop. Points}\\
       &  \textbf{State Acc.} & \textbf{State Acc.} &  \textbf{Accuracy} & \textbf{Points} & \textbf{Followed} & \textbf{Scored} \\ \hline \hline
       \multicolumn{7}{|l|}{\textbf{Development Results \& Ablation Analysis}} \\ \hline
        Full model &  \stdev{58.2}{0.5} & \stdev{32.6}{0.8} & \stdev{15.8}{0.5} & \stdev{0.66}{0.1} & \stdev{20.5}{1.2} & \stdev{18.1}{0.8} \\ \dline
       -- Trajectory distribution &  \stdev{38.5}{2.7} & \stdev{10.1}{2.7} & \stdev{\phantom{0}5.5}{2.6} & \stdev{0.29}{0.02} & \stdev{10.0}{0.9} & \stdev{\phantom{0}7.9}{0.7} \\
       -- \textsc{GOAL} distribution &  \stdev{56.2}{1.5} & \stdev{30.8}{0.4} & \stdev{14.9}{0.3} & \stdev{0.66}{0.09} & \stdev{17.9}{1.0} & \stdev{15.9}{1.3} \\ 
       -- \textsc{AVOID} distribution &  \stdev{57.0}{0.3} & \stdev{32.6}{1.6} & \stdev{15.4}{1.3} & \stdev{0.63}{0.04} & \stdev{18.8}{1.5} & \stdev{17.8}{0.7} \\
       -- \textsc{NOPASS} distribution & \stdev{59.2}{0.5} & \stdev{32.0}{0.8} & \stdev{15.0}{0.5} & \stdev{0.70}{0.03} & \stdev{18.4}{0.9} & \stdev{16.6}{0.9} \\
       -- Action recurrence &  \stdev{42.3}{1.5} &  \stdev{16.7}{1.2} & \stdev{10.0}{0.7} & \stdev{0.42}{0.03} & \stdev{12.8}{1.7} & \stdev{10.7}{0.5} \\ 
       -- Fine-tuning &   \stdev{43.6}{1.9} & \stdev{\phantom{0}8.5}{1.1} & \stdev{\phantom{0}4.5}{0.5} & \stdev{0.65}{0.09} &  \stdev{14.1}{1.3} & \stdev{\phantom{0}9.2}{0.9} \\
       -- Early goal auxiliary & \stdev{57.2}{2.3} & \stdev{31.2}{1.7} & \stdev{14.9}{1.6} & \stdev{0.65}{0.05} & \stdev{17.9}{1.1} & \stdev{16.5}{0.7} \\ 
       -- Example aggregation &  \stdev{59.4}{1.8}  & \stdev{32.0}{1.0} & \stdev{15.7}{0.6} & \stdev{0.65}{0.09} & \stdev{20.4}{1.4} & \stdev{16.5}{0.4}\\ 
        -- Implicit discriminator  &  \stdev{57.5}{2.1} & \stdev{32.7}{1.0} & \stdev{16.4}{0.3} & \stdev{0.70}{0.02} & \stdev{18.8}{1.8}& \stdev{16.7}{0.6} \\ \dline
       -- Instructions &  \stdev{15.5}{1.5} & \stdev{\phantom{0}2.7}{1.5} & \stdev{\phantom{0}1.2}{1.2} & \stdev{0.24}{0.07} & \stdev{\phantom{0}4.4}{1.0} & \stdev{\phantom{0}4.6}{0.7} \\ \dline
       + Gold plan &  \stdev{87.4}{0.5} & \stdev{80.2}{0.2} & \stdev{63.4}{0.2} & -- & --  & --\\ \dline
      \system{Seq2seq+attn} &  \stdev{35.3}{0.8} & \stdev{11.1}{0.5}& \stdev{\phantom{0}9.4}{0.5} & \stdev{0.20}{0.04} & \stdev{\phantom{0}8.8}{0.1}& \stdev{\phantom{0}6.3}{0.1}  \\ \dline
      Static oracle &  99.7 & 99.7 & 100.0  & 6.58 & 98.5 & 97.9 \\
      \hline \hline
    \multicolumn{7}{|l|}{\textbf{Test Results}} \\ \hline
    Full model &  58.4 & 32.1 & 15.6 & 0.62 & 15.4 & 17.9 \\ \dline
      \system{Seq2seq+attn}  & 37.3 & 10.8 & \phantom{0}8.5 & 0.22 & \phantom{0}8.7 & \phantom{0}6.5  \\ \dline
     Static oracle &  99.7 & 99.7 & 100.0 & 6.66 & 96.8 & 95.6 \\ \hline
    \end{tabular}
    \caption{Development and test results on all systems, including ablation results.}
    \label{tab:static_results}
\end{table*}

\paragraph{Systems}
We evaluate three systems: (a) the full model; (b) \system{Seq2seq+attn}:\footnote{This baseline is similar to  \citet{Mei:16neuralnavi}.} sequence-to-sequence with attention; and (c) a static oracle that executes the gold sequence of actions in the recorded interaction.
We report mean and standard deviation across three  trials for development results. 
We ablate  model and learning components, and additionally evaluate the action generator with access to gold plans.\footnote{We do not measure interaction-level metrics with gold plans as they are only available for the gold start positions.}
On the test set and for human evaluation, we use the model with the highest proportion of points scored.
We provide implementation and learning  details in Appendix~\ref{sec:sup:impl}.

\section{Results}
\label{sec:results}

Table~\ref{tab:static_results} shows development and test results, including ablations.
We consider the proportion of points scored computed with cascaded evaluation as the main metric.
Our complete approach significantly outperforms \system{Seq2seq+attn}. 
Key to this difference is the added structure within the model and the direct supervision on it. 
The results also show the large remaining gap to the static oracle.\footnote{Appendix~\ref{sec:sup:cascaded} explains the static oracle performance.} 

Our results show how considering error propagation for all available instructions in cascaded evaluation guides different design choices. 
For example, example aggregation and the implicit discriminator lower performance according to instruction-level metrics, which do not consider error propagation. We see a similar trend for the implicit discriminator when  looking at full game points, an interaction-level metric that does not account for performance on over 80\% of the data because of error propagation. 
In contrast, the proportion of points scored computed using cascaded evaluation shows the benefit of both mechanisms. 

Our ablations demonstrate the benefit of each model component. 
All  four distributions help.
Without the trajectory distribution (-- Trajectory distribution), performance drops almost to the level of \system{Seq2seq+attn}. This indicates the action predictor is not robust enough to construct a path given only the three other disjoint distributions. 
While the predicted trajectory distribution contains all information necessary to reach the correct cards and goal location, the other three distributions further improve performance.
This is likely because redundancy with the trajectory distribution makes the model more robust to noisy predictions in the trajectory distribution.
For example, the GOAL distribution guides the agent to move towards goal cards even if the predicted trajectory is discontinuous.
The action generation recurrence is also critical (-- Action recurrence), allowing the agent to keep track of which locations it already passed when navigating complex paths that branch, loop, or overlap with themselves.

While we observe that each stage separately performs well after pretraining, combining them without fine-tuning (-- Fine-tuning) leads to low performance because of the shift in the second stage input. 
Providing the gold distributions to the action generator illustrates this (+ Gold plan). 
Removing early goal auxiliary loss $\mathcal{L}_{G'}$ (Section~\ref{sec:learn:pretrain}) leads to a slight drop in performance on all metrics (-- Early goal auxiliary).
Learning with aggregated recovery examples helps the model to learn to recover from errors in previous instructions and increases the proportion of points scored (-- Example aggregation). However, without the implicit reasoning discriminator (\mbox{-- Implicit discriminator}), the additional examples make learning too difficult, and do not help. 
Finally, removing the  language input (-- Instructions)  significantly decreases performance, showing that the data is relatively robust to observational biases and language is necessary for the task.

In the human evaluation, we observe a mean of 6.2 points (max of 14) with our follower model, compared to 12.7 (max of 20) with human followers. While this shows there is much room for improvement, it illustrates how human leaders adapt and use the agent effectively. 
One key strategy of adaptation is to use simplified language that fits the model better. This includes shorter instructions, with 8.5 tokens on average with automated followers compared  to 12.3 with humans, and a smaller vocabulary, 578 word types with automated followers and 1037 with humans. 
In general, human leaders commented that they are able to easily distinguish between automated and human followers, and find working with the automated agent frustrating.

\section{Discussion}
\label{sec:discussion}

Our human evaluation highlights several directions for future work. 
While human leaders adapt to the agent, scoring up to 14 points, there remains a significant gap to collaborations with human followers. 
Reported errors include getting stuck behind objects, selecting unmentioned cards, going in the wrong direction, and ignoring instructions. 
At least one worker developed a strategy that took advantage of the agent's full observability, writing instructions with only simple card references. 
An important direction for future work is to remove our full observability assumption. 
Other future directions include experimenting with using the interaction history, expanding the learning example aggregation to error cases beyond incorrect start positions, and  making agent reasoning interpretable to reduce user frustration. 
\thegame also provides opportunities to study pragmatic reasoning for language understanding~\cite{Andreas:16pragmatics,Fried:17pragmatic-models,Liang:19collaboration}.
While we currently focus on language understanding by limiting the communication to be unidirectional, bidirectional communication would allow for more natural and efficient collaborations~\cite{Potts:12,Ilinykh:19meetup}.
\thegame could be easily adapted to allow bidirectional communication, and provide a platform to study challenges in language generation.

\section*{Acknowledgments}

This research was supported by the NSF under Grant No. 1750499, a Google Focused Award, an AI2 KSC Award, a Workday Faculty Award, Unity, and an Amazon Cloud Credits Grant. 
This material is based on work supported by the National Science Foundation Graduate Research Fellowship under Grant No. DGE-1650441. 
We thank Valts Blukis, Jin Sun, and Mark Yatskar for comments and suggestions, the workers who participated in our data collection, and the  reviewers.

\bibliography{main}
\bibliographystyle{acl_natbib}

\clearpage

\appendix
\appendix

\section{\thegame Game Design}
\label{sec:sup:game}

This appendix supplements Section~\ref{sec:overview} with further game design details and discussion of the reasoning behind them. 

\paragraph{World View}

Figure~\ref{fig:interface_leader} shows the leader's point of view, and Figure~\ref{fig:interface_follower} shows the follower's. 
The leader observes the entire environment, while the follower only has access to a restricted first person view. 
The leader can also toggle to an overhead view to see obstructed cards using the camera button, and has access to the follower's current view to aide them in writing instructions that make sense to the follower.
Selected cards are outlined in blue for both players.
Invalid selections appear in red for the leader only. 
This setup makes the follower dependent on the leader, limits the follower ability to plan the card collection strategy, and encourages  collaboration.

\paragraph{Game Progression}

The two players switch control of the game by taking turns.
During each turn, the follower can take ten ($\maxsteps_f = 10$) steps while the leader can take five ($\maxsteps_l = 5$). 
Allowing the follower more steps than the leader incentivizes  delegating lengthier tasks to the follower, such as grabbing multiple cards per turn or moving further away. 
We do not count actions which do not change the player's location or rotation, such as moving forward into an obstacle, against this limit.
We additionally limit the amount of time each player has per turn. 
This requires players to move quickly without frustrating their partner by taking a long time, and additionally limits the maximum time per game.
Both players begin with six turns each.
The game ends when the players run out of turns.

The leader turn ends once they press the end turn button or after 45 seconds. The end turn button is disabled as long as there are no instructions in the follower queue to nudge the leader to use the follower if time allows it. 
The allotted 45 seconds allow the leader sufficient time to move, plan, and write instructions. 
During the leader's turn, they can add any number of new instructions to the queue.

The follower only receives control if there are instructions in the queue. 
If the queue is empty when the leader finishes their turn, the follower's turn is skipped, but the number of turns remaining still decreases.
The follower's turn ends automatically when they run out of steps, after 15 seconds, or when they complete all instructions in the queue. 
During the follower's turn, they can mark any number of instructions as complete using the $\doneaction$ action.
The follower sees the current and previous instructions, even if there are more instructions in the queue. 
They must mark the current instruction as complete before seeing the next.
This is done to simplify the reasoning available to the follower. For example, to avoid cases where the follower skips a command based on future ones. 
Because there may be more future instructions in the queue, this incentivizes the follower to not waste moves in the current instruction and be as efficient as possible.
During data collection, this provides alignment of actions to instructions because it prohibits a follower from taking actions aligning with a future instruction without marking the current instruction as complete. 
Without instruction completion annotation, the problem of alignment between instructions and actions becomes much more difficult when processing the recorded interactions.

\paragraph{Scoring Points}
When a valid set is made, the selected cards disappear, and three cards are randomly generated and placed on the grid such that the new grid contains a least one valid set.
The two players earn a point, and are given extra turns.
The number of added turns decays as they complete more sets, eventually reaching zero added turns.
The maximum possible number of turns in a game is 65. In the training data, 454 games reached this number of turns. 
Adding extra turns when a set is made allows us to collect more data from games that are going well. 
It also allows us to pay players based on the number of sets completed, and incentivizes them to play as well as possible. 
If a game is going poorly, e.g., if the pair fails to earn a point in the first six turns, the game will end early. 
However, if the game is going well, implying the pair is collaborating well, the game will continue for longer, and will contain a longer sequence of instructions.

\section{\thegame Transition Function}
\label{sec:sup:transfunc}

\begin{table*}[t]
    \centering\footnotesize
    \begin{tabular}{|c|r|rcl|} \hline
    \textbf{Rule No. } &  \multicolumn{1}{|c|}{\textbf{Domain}} & \multicolumn{3}{|c|}{\textbf{Definition}} \\ \hline
        1 &  $\forall \sentence \in \sentences$, $\state \in \states$ & $\transfunc(\state, \langle \queue, \leader, \stepsremaining \rangle, \sentence)$ & $=$ & $(\state, \langle \queue \sentence, \leader, \stepsremaining \rangle)$\\ \hline
        2 &  $\forall \state \in \states$, $|\queue| \geq 1$ & $\transfunc(\state, \langle \queue, \leader, \stepsremaining \rangle, \doneaction)$  & $=$ & $(\state, \langle \queue, \follower, \maxsteps_f \rangle)$\\ \hline
     3 & $\forall \state \in \states$, $|\queue| = 0$ & $\transfunc(\state, \langle \langle~\rangle, \leader, \stepsremaining \rangle, \doneaction)$  & $=$ & $(\state \langle \queue, \leader, \maxsteps_l \rangle)$\\ \hline 
 4  & $\forall \action \in \worldactions$, $\state \in \states$ & $\transfunc(\state, \langle \queue, \leader, 1 \rangle, \action)$ & $=$ & $(\transfunc_w(\state, \action), \langle \queue, \leader, 0 \rangle)$\\ \hline \hline
   5  & $\forall \state \in \states$, $|\queue| > 1$ & $\transfunc(\state, \langle \sentence\queue, \follower, \stepsremaining \rangle, \doneaction)$ & $=$ & $(\state, \langle \queue, \follower, \stepsremaining\rangle )$ \\ \hline
   6  & $\forall \state \in \states$, $|\queue| = 1$ & $\transfunc(\state, \langle \queue, \follower, \stepsremaining \rangle, \doneaction)$ & $=$ & $(\state, \langle \langle~\rangle, \leader, \maxsteps_l \rangle )$ \\ \hline
  7 & $\forall \action \in \worldactions$, $\state \in \states$ & $\transfunc(\state, \langle \queue, \follower, 1\rangle, \action)$ & $=$ & $(\transfunc_w(\state, \action), \langle \queue, \leader, \maxsteps_l\rangle)$\\ \hline \hline
  \multirow{3}{*}{8} & $\forall \action \in \worldactions$, $\state \in \states$ & \multirow{3}{*}{$\transfunc\left(\state, \langle \queue, \turntaker, \stepsremaining \rangle, \action\right)$} & \multirow{3}{*}{$=$} & \multirow{3}{*}{$\left(\transfunc_w(\state, \action), \langle \queue, \turntaker, \stepsremaining - 1\rangle\right)$}   \\
  & $\forall\stepsremaining \in \naturals_{>1}$ & & & \\ 
   & $\forall \turntaker \in \{ \leader, \follower\}$ & & & \\ \hline
    \end{tabular}
    \caption{Definition of transition function $\transfunc$. $\transfunc_w$ is the world state transition function.}
    \label{tab:transition_func}
\end{table*}

The transition function in  \thegame $\transfunc: \states \times \istates \times \actions \rightarrow \states \times \istates$ is formally defined in Table~\ref{tab:transition_func}.
Each of the rules in the table is additionally associated with a domain over which it is not defined, for example when $\turntaker = \follower$ and $\action \in \sentences$ (i.e., the follower can not give instructions). The rules are:

\begin{enumerate}[start=1,label={\bfseries Rule \arabic*:},labelwidth=3em, itemindent=2em]
  \item When an instruction is issued, it is added to the end  of the queue. This action does not use a step, so the number of steps remaining $\stepsremaining$ does not decrease. This rule is not defined when $\turntaker = \follower$ because the follower cannot give an instruction. 
  \item When the leader ends their turn, and the queue is not empty, control switches to the follower, and the number of steps remaining in the turn is the maximum number for the follower $\maxsteps_f$.
  \item When the leader ends their turn, and the queue is empty, control does not switch to the follower; instead, a new leader turn begins with $\maxsteps_l$ available steps.
  \item When the leader runs out of remaining steps, control does not immediately switch to the follower. This allows the leader to issue more instructions before manually ending their turn or when their time runs out. 
  \item When the follower marks an instruction as finished, and more instructions  remain in the queue, the current instruction at the head of the queue is removed. This action does not use a step.
  \item When the follower marks an instruction as finished, if the finished instruction was the last in the queue, control automatically switches to the leader with $\maxsteps_l$ remaining steps. 
  \item When the follower runs out of steps in their turn, control immediately switches to the leader with $\maxsteps_l$ remaining steps.
  \item Both agents can take actions which modify the world state $\state$. Each such action $\action \in \worldactions$ costs a step. We assume access to a domain-specific transition function, $\transfunc_w : \states \times \worldactions \rightarrow \states$, that describes how an environment action modifies the environment. There may exist combinations of states and actions for which $\transfunc_w$ is not defined; for example, an agent moving forward onto an obstacle. Additionally, $\forall \state \in \states$ and $ \action \in \worldactions$, $\transfunc(\state, \langle Q, \leader, 0 \rangle, \action )$ results in an invalid state because, while the leader can still issue instructions after running out of steps, they cannot move.
\end{enumerate}

\section{Data Collection Details}
\label{sec:sup:data}

\begin{figure*}
    \centering
    \fbox{\includegraphics[width=\textwidth]{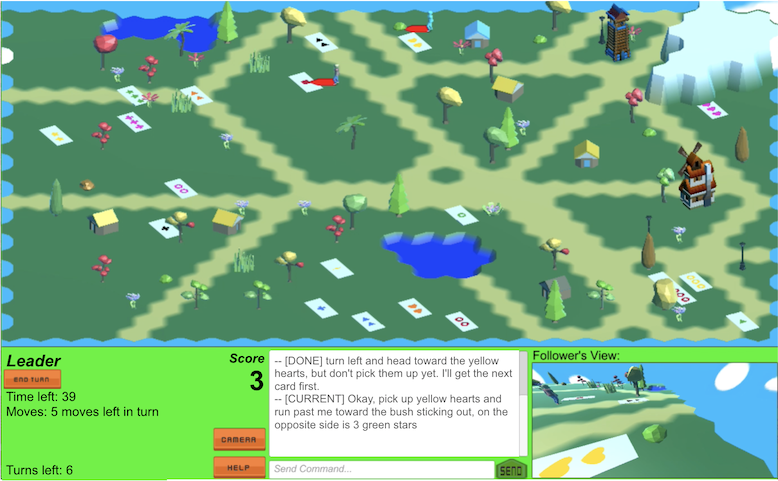}}
    \caption{The \thegame leader gameplay interface.}
    \label{fig:interface_leader}
\end{figure*}

\begin{figure*}
    \centering
    \fbox{\includegraphics[width=\linewidth]{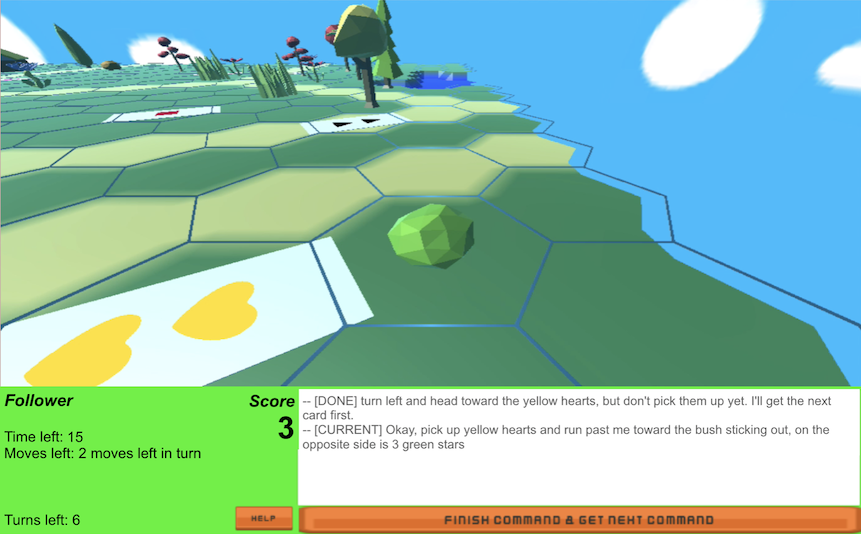}}
    \caption{The \thegame follower gameplay interface.}
    \label{fig:interface_follower}
\end{figure*}

Figures~\ref{fig:interface_leader} and~\ref{fig:interface_follower} show the leader's and follower's interfaces.

\paragraph{Crowdsourcing Management}

We use a qualification task to both teach workers how to play the game and to mark workers as qualified for our main task.
We restrict those who can qualify to workers located in majority English-speaking countries with at least 90\% approved HITs and at least 100 completed HITs.
The qualification task has three components: an interactive tutorial for the leader role, an interactive  tutorial for the follower role, and a short quiz about the gameplay.
In both tutorials, turn-switching is disabled and workers have an unlimited number of moves to use to complete the tutorial.
Each tutorial uses the same map. This allows us to pre-program instructions for the tutorials. 

In the leader tutorial, the worker has access to the full game board. 
They are asked to send a command to the follower, and are instructed via in-game prompts to collect a specific set of cards.
Finally, they are asked to collect two more sets in the environment that are valid. 
Workers who send a command and collect a total of three sets successfully complete this tutorial.

In the follower tutorial, the worker has access only to the follower view.
Pre-written commands are issued to the worker, and they must follow them one-by-one to complete a set.
The commands include an example of the leader correcting a set-planning mistake.
If the worker marks all commands as finished and successfully collects one set, the follower tutorial is complete.

Finally, workers are asked to read the game instructions and complete a short quiz.
They are asked questions regarding the validity of card sets, the responsibilities of both players, and how each game ends.

We maintain two groups of workers split by experience with the game, and use separate pools of HITs for each.
A worker can join the expert pool if they have shown they understand how to play as a leader and as a follower through at least one game each.
This allows new players to learn the game rules without frustrating expert players.
At the end of data collection, 95 workers were in the expert pool while 169 were in the non-expert pool, for a total of 264 participating workers.

We pay workers a bonus per point they earn, increasing the bonus as more points are earned, in addition to a base pay of per game.
We do not pay leaders and followers differently.
The median game cost was \$5.80.

\paragraph{The \thegame Dataset}

In total, we collect 1,526 games played by both experts and non-experts. 
Of these, we keep 1,202 (78.8\%) games, comprising 23,979 total instructions, discarding those where no instructions were complete, or where alignment between instructions and actions was suspected low-quality.
For example, we removed interactions with a low proportion of instructions being marked as complete, or very long action sequences from the follower, both which indicate the follower did not properly complete instructions.

When splitting the data, we ensured the mean score between the three splits was roughly the same. 
Table~\ref{tab:dataset_statistics} shows basic statistics of the data we collected after pruning. 
82.6\% of post-pruning games are from the expert pool.
In the training set, the mean number of completed instructions is 19.9 and the median is 24.0. 
83.3\% of games have a score greater than zero. We include games with a score of zero if the alignment between instructions and actions is high-quality according to our pruning heuristics. 
The vocabulary size is computed by lowercasing all word types and tokenizing using the NLTK word tokenizer. 
Our dataset contains longer interactions than several existing datasets for sequential instruction following and interaction~\cite[e.g.,][]{Chen:11,Long:16context,He:17dialogue,Devries:18,Kim:19codraw,Hu:19minirts,Udagawa:19}, though still shorter than the Cards corpus~\cite{Djalali11:cards-qud,Djalali11:cards-preference,Potts:12}.
Individual sentences are also longer than several similar corpora~\cite[e.g.,][]{Chen:11,Djalali11:cards-qud,Long:16context,He:17dialogue,Hu:19minirts}.

\begin{table}[t] 
  \footnotesize
    \centering
    \begin{tabular}{|c|c|c|c|} \hline
        & \textbf{Mean} & \textbf{Median} & \textbf{Max}  \\ \hline
    {Total Game Time (m:s)} & 16:28 & 18:40 & 31:31\\ \hline
    {Score / Interaction} & 7.9 & 9.0 & 19 \\ \hline
    {\# Instr. / Interaction} & 19.9 & 24.0 & 40 \\ \hline
    {\# Tokens / Instr.} & 14.0 & 13.0 & 55 \\ \hline
    {\# Follower Actions / Instr.} & 8.5 & 8.0 & 50 \\ \hline \hline
    {\# Interactions} & \multicolumn{3}{|c|}{1,202} \\ \hline
    {Vocabulary Size} & \multicolumn{3}{|c|}{3,641} \\ \hline
    \end{tabular}
    \caption{Human-human games data statistics. All statistics except the number of examples are computed on the training set only.}
    \label{tab:dataset_statistics}
\end{table}

\section{Model Architecture Details}
\label{app:sec:lingunet}

\paragraph{\lingunet Formal Description}

We provide a formal description of \system{LingUNet} for reference only. 
\system{LingUNet} was originally introduced by \citet{Misra:18goalprediction} and \citet{Blukis:18drone}.

The input to $\lingunet$ are the environment representation $\mathbf{F}_0$ and instruction representation $\mathbf{\sentence}$. 
$\lingunet$ consists of three major stages: a series of convolutions on $\mathbf{F}_0$, a series of text-based convolutions derived from $\mathbf{\sentence}$, and a series of transposed convolutions to form a final prediction.
The output of the $\lingunet$ is a feature map with the same width and height as $\mathbf{F}_0$. 
Each stage has the same number of operations, which we refer to as the depth $\lingunetdepth$.

First, a series of $\lingunetdepth$ convolutional layers is applied to  $\mathbf{F}_0$.
Each layer at depth $l$ is a sequence of two convolution operations separated by a leaky ReLU non-linearity: 

\begin{small}
\begin{equation*}
    \mathbf{F}_l = \textsc{Norm}(\textsc{ReLU}(\textsc{ReLU}(\mathbf{F}_{l-1} * \mathbf{K}^C_l)* \mathbf{K}^{C \prime}_l )) \;\; .
\end{equation*}
\end{small}

\noindent We use a stride of two when convolving with $\mathbf{K}^{C\prime}_l$, and do not apply  $\textsc{Norm}$  when $l = \lingunetdepth$.

In the second stage, the instruction representation $\mathbf{\sentence}$ is split into $\lingunetdepth$ segments $\mathbf{\sentence}_l$ such that $\mathbf{\sentence} = [ \mathbf{\sentence}_1; \dots; \mathbf{\sentence}_\lingunetdepth]$ and segments have equal length.
Each segment is mapped to a $1 \times 1$ kernel $\mathbf{K}^I_l$ using learned weights $\mathbf{W}^I_l$ and biases $\mathbf{b}^I_l$.
$\mathbf{K}^I_l$ is normalized and used to convolve over $\mathbf{F}_l$:

\begin{small}
\begin{equation*}
    \mathbf{G}_l = \textsc{Norm}(\mathbf{F}_l * ||\mathbf{K}^I_l||_2) \;\; .
\end{equation*}
\end{small}

\noindent As before, we do not apply $\textsc{Norm}$  when $l = \lingunetdepth$.

In the last stage, a series of transposed convolutions\footnote{We use $*^\top$ to represent the transposed convolution operation.} are applied starting from the bottom layer  and gradually synthesizing a larger feature map. For $l > 1$:

\begin{small}
\begin{equation*}
    \mathbf{H}_l = \textsc{Norm}(\textsc{ReLU}(\textsc{ReLU}([\mathbf{H}_{l+1}; \mathbf{G}_l] {*^\top} \mathbf{K}^T_l) *^\top \mathbf{K}^{T\prime}_{l})) \;\;,
\end{equation*}
\end{small}

\noindent
where $[\mathbf{H}; \mathbf{G}]$ indicates channel-wise concatenation of feature maps $\mathbf{H}$ and $\mathbf{G}$, $\mathbf{H}_{H+1}$ is a zero matrix, and $\textsc{Norm}$ is not applied when $l = \lingunetdepth$.
We use a stride of two when convolving with $\mathbf{K}^{T\prime}_l$.
At the topmost layer of $\lingunet$, a final transposed convolution is applied to form a feature map $\mathbf{H}'_1$: 

\begin{small}
\begin{equation*}
    \mathbf{H}'_1 = [\mathbf{H}_2; \mathbf{G}_1] *^\top \mathbf{K}^T_1 \;\; .
\end{equation*}
\end{small}

The top layer $\mathbf{H}'_1 \in \mathbb{R}^{4 \times W \times H}$ is split into the four planning distributions as the output of the $\lingunet$.

\paragraph{Frames of Reference}

The world state is first embedded using a feature lookup and a text-conditioned kernel (Section~\ref{sec:model}; Input Representation).
This feature map is rotated and centered to create $\mathbf{F}_0$, so that the agent's location when beginning to follow the instruction is in the center, and the agent is facing in a consistent direction.
Therefore,  $\lingunet$ (Section~\ref{sec:model}; Stage 1: Plan Prediction) operates over a feature map relative to the agent's frame of reference at the time of starting to follow the instruction.

The action generator (Section~\ref{sec:model}; Stage 2: Action Generation) also operates on feature maps relative to the agent's frame of reference, updated as the agent moves and turns in the environment  changing its location and orientation.
At each action generation prediction step, the concatenated planning distributions $\mathbf{P}$ are rotated, centered, and cropped around the agent's current orientation. This orientation is determined by the orientation when starting the instructions and the actions it has executed so far for the current instruction.

\section{Learning Details}

\subsection{Stage 1 Loss Computation}
\label{sec:sup:learn}

This section provides formal details of the loss computation used in Section~\ref{sec:learn:pretrain}. 
For ease of notation, we consider a single example $\interaction = \langle (\state_1, \istate_1, \action_1), \dots, (\state_n, \istate_n, \action_n)\rangle$, where the instruction at the head of the queue $\queue$ is $\sentence$.

The loss of the visitation distribution $p(\position \mid \state_1, \sentence)$ is:

\begin{small}
\begin{equation*}
    \mathcal{L}_V(\param_1) = -\sum_\position p^*_V(\position) \log p(\position \mid  \state_1, \sentence)\;\;,
\end{equation*}
\end{small}

\noindent
where the summation is over all positions $\position$ in the environment and $p^*_V(\position)$ is proportional to the number of states $s_t \in \interaction$ where the follower is in position $\position$.

We compute the goal and avoidance distribution losses only for positions that have cards:

\begin{small}
\begin{eqnarray*}
    \mathcal{L}_G (\param_1) =& \\ &\hspace{-1cm} -\frac{1}{W\times H}\sum_{\position \in C} p^*_G(\position) \log p(\mathrm{GOAL} = 1 \mid \position, \state_1, \sentence) \\
    \mathcal{L}_A(\param_1) =& \\ &\hspace{-1cm} -\frac{1}{W\times H}\sum_{\position \in C} p^*_A(\position) \log p(\mathrm{AVOID} = 1 \mid \position, \state_1, \sentence)\;\;,
\end{eqnarray*}
\end{small}

\noindent 
where $C$ is the set positions that contain cards, $W$ is the width of the environment, and $H$ is the height. 
We set $p^*_G(\position)$ to 1 for all $\position$ that contain a card that the follower changed its selection status in $\interaction$, and 0 for all other positions. 
Similarly, we set $p^*_A(\position)$ to 1 for all $\position$ that have cards that the follower does not change during the interaction $\interaction$, but zero for the initial position regardless of whether it contains a card.

The loss for the no passing distribution is:  

\begin{small}
\begin{eqnarray*}
    \mathcal{L}_P(\param_1) =& \\ &\hspace{-1cm} -\frac{1}{W\times H}\sum_{\position} p^*_P(\position)\log p(\mathrm{NOPASS} = 1\mid \position, \state_1, \sentence)\;\;,
\end{eqnarray*}
\end{small}

\noindent 
where $p^*_P(\position)$ is 1 for all positions the agent cannot  move onto, and zero otherwise.

The auxiliary  goal-prediction loss is:

\begin{small}
\begin{eqnarray*}
    \mathcal{L}_{G'}(\param_1) =& \\ &\hspace{-1cm} -\frac{1}{W\times H}\sum_{\position \in C} p^*_G(\position) \log p'_G(\mathrm{GOAL} = 1 \mid \position, \state_1, \sentence)\;\;.
\end{eqnarray*}
\end{small}

\noindent
We compute the goal probability with the learned parameters $\mathbf{W}^{G'}$ and $\mathbf{b}^{G'}$:

\begin{small}
\begin{equation*}
  p'_G(\mathrm{GOAL} = 1 \mid \position, \state_1, \sentence ) = \sigma(\mathbf{W}^{G'} \envembedding'_\position  + \mathbf{b}^{G'})\;\;,
\end{equation*}  
\end{small}

\noindent
where $\envembedding'_\position$ is the vector along the channel dimension for position $\position$ in the environment embedding tensor $\envembedding'$.

\subsection{Example Aggregation}
\label{sec:app:learning}

\paragraph{Error Classes}

We identify two classes of  erroneous states in \thegame: 
(a) not selecting the correct set of cards specified by the instruction; 
and (b) finishing with the right card selection, but stopping at the wrong position. 
To recover from case (a), the agent could unselect cards it shouldn't have selected, or select cards it missed.
Alternatively, the agent could recognize it has made an error, and instead stop and wait for the next leader instruction, anticipating a correction.
However, learning this requires access to previous world states and instructions.
We focus on modification of the learning algorithm using example aggregation, and leave this case for future work. 
We instead target class (b), and add a discriminator to the model to allow the model to learn different reasoning for examples that require implicit actions, as discussed in Section~\ref{sec:learn:finetune}.

\paragraph{Creating Recovery Examples}

The oracle generates a sequence of state-action pairs to go from $\state'$, the incorrect initial state from the previous instruction, to state $\state_t$ at index $t$ in the correct sequence such that $\state_t$ is either the first state in the sequence where a card's state changes, or if no cards are changed, the final state $s_n$.
The oracle finds a sequence of state-action pairs expressing the shortest path $\state'$ to $\state_t$. 
Finally, it appends the remainder of the correct state-action sequence starting from index $t$, $\langle (\state_t, \istate_t, \action_t), \dots, (\state_n, \istate_n, \action_n) \rangle$.

If the correct sequence for $\interaction^{(i, {j+1})}$ is $\langle \state_n, \istate_n, \doneaction \rangle$ (i.e., no action was done in the original example), we do not generate a new path, but instead use the state-action sequence $\langle \state', \istate', \doneaction \rangle$ as annotation for $\interaction^{\prime(i,j+1)}$. 
These examples are annotated as not requiring implicit reasoning.

During inference on the previous example $\interaction^{(i, j)}$, it is possible that some leader actions associated with that example may not be executed (i.e., if the follower predicted $\doneaction$ too soon).
If this happens, the leader must execute actions to `catch up' to the follower in the generated recovery example. 
We first find the sequence of leader actions starting from the first leader turn associated with $\interaction^{(i, j)}$ that was not executed during inference, to the final leader turn associated with $\interaction^{(i, {j+1})}$.
When generating the recovery sequence $\interaction^{\prime(i,j+1)}$, we take into consideration this sequence as affecting the world states $\state$.
For example, suppose that the agent stops a turn early during inference, and the final leader's turn consisting of actions $\langle \mathtt{FORWARD}, \mathtt{FORWARD}, \mathtt{FORWARD}, \doneaction \rangle$ was not executed.
Instead of stopping in, for example, position $(3, 0)$, this may mean the leader has stopped in position $(0, 0)$.
When creating the recovery example, the first world state $\state_0$ shows the leader at position $(0, 0)$ rather than $(3, 0)$. To correct this, the recovery example will start with a leader turn, where the leader executes the sequence $\langle \mathtt{FORWARD}, \mathtt{FORWARD}, \mathtt{FORWARD}, \doneaction \rangle$.

\section{Evaluation Details}
\label{sec:sup:cascaded}

\paragraph{Cascaded Evaluation}
To compute metrics using cascaded evaluation, we construct a set of cascaded evaluation examples from the original test set.
We assume access to a test set of $M$ recorded interactions $\left\{ \interaction^{(i)} \right\}^M_{i=1}$, where each $\interaction^{(i)} = \left\langle \left(\state_1^{(i)}, \istate_1^{(i)}, \action_1^{(i)}\right), \dots, \left(\state_{|\interaction|}^{(i)}, \istate_{|\interaction|}^{(i)}, \action_{|\interaction|}^{(i)}\right) \right\rangle$. 
For each instruction $\sentence_j$ in $\interaction^{(i)}$, we create an example $\interaction_C^{(i,j)} = \left\langle \left(\state_{j'}^{(i)}, \istate_{j'}^{(i)}, \action_{j'}^{(i)}\right), \dots, \left(\state_{|\interaction|}^{(i)}, \istate_{|\interaction|}^{(i)}, \action_{|\interaction|}^{(i)}\right) \right\rangle$, where $j'$ is the first follower step of executing $\sentence_j$. 
We treat each $\interaction_C^{(i,j)}$ as a separate example. 
For each metric, we report the proportion of the maximum value possible for each $\interaction_C^{(i,j)}$, and average across all examples $\interaction_C^{(i,j)}$. 
When computing the proportion of instructions followed in cascaded evaluation, the maximum possible for example  $\interaction_C^{(i,j)}$ is the number of remaining instructions $N-j$ where $N$ is the number of instructions in $\interaction^{(i)}$. 
When computing the proportion of points scored, we subtract the points scored in the game before step $j$ to only account for points possible in the instructions present in $\interaction_C^{(i,j)}$. 

\paragraph{Performance of the Static Oracle}
The static oracle does not have perfect performance. 
This is because the follower's turn ended before all ten steps were used  in some recorded interactions. 
During evaluation, however, we allow the follower to move for all ten available steps.
This sometimes leads to misalignment between leader and follower actions. This means some expected sets can not be completed.

\section{Implementation and Hyperparameters}
\label{sec:sup:impl}

\paragraph{Hyperparameters}
We tune hyperparameters on the development set.
We use a word embedding size of 64, and encode instructions into a vector of length 64 using a single-layer RNN with LSTM units.
We lowercase words in the vocabulary and map all words with a frequency of one in the training set to a single out-of-vocabulary token.
We use a hex property embedding size of 32. 
$\mathbf{S}'$ has four channels.
The text-based kernels map to a feature map with 24 channels.
The convolution and transpose convolution phases of $\lingunet$ use kernel sizes of three.

The action generator uses a forward RNN with a single layer consisting of 128 LSTM hidden units.
The action embedding size is 32.
We rotate, transform, and crop the input plan distribution to a $4 \times 5 \times 5$ feature map around the agent's current position and rotation for each generated action.
$\textsc{CNN}^P$ maps the cropped distributions to a feature map with eight channels, and has a kernel size of three and stride of one.
During fine-tuning, each $\mathbf{K}^\mathrm{IMP}_l$ does not have biases.
For all LSTMs,  we initialize the hidden state $\mathbf{h}_0$ as a zero vector. 
For brevity, cell memory $\mathbf{c}^D$, also initialized as a zero vector, is omitted from  RNN descriptions.

\paragraph{Learning}
The plan prediction stage (Stage 1) includes the following parameters and parameterized components: $\embed^{\sentences}$, $\rnn^\sentences$, $\embed^{\states}$, $\mathbf{W}_s$, $\mathbf{b}_s$, and $\lingunet$. 
The action generation stage (Stage 2) includes the following parameters and parameterized components: $\textsc{CNN}^P$, $\mathbf{W}^P_1$, $\mathbf{W}^P_2$, $\mathbf{b}^P_1$, $\mathbf{b}^P_2$, $\embed^{\actions}$, $\rnn^\actions$, $\mathbf{W}^\actions$, and $\mathbf{b}^\actions$. 
We add the following parameters for the early goal prediction auxiliary objective and implicit reasoning discriminator $\mathbf{W}^{G'}$, $\mathbf{b}^{G'}$, and $\kernel^{\rm IMP}_l$, $1<l<L$. 

For pretraining Stage 1, we use a learning rate of 0.0075 using  \textsc{Adam}~\cite{Kingma:14adam} and an L2 coefficient of $10^-6$.
For pretraining Stage 2 and during fine-tuning, we use a learning rate of 0.001 and  \textsc{Adam} with no L2 regularization.
For pretraining Stage 1 and during fine-tuning, $\lambda_V = 0.04$, $\lambda_G = 1$, $\lambda_A = 0.1$, $\lambda_P = 0.1$, and $\lambda_{G'} = 1$.
During fine-tuning, $\lambda_\mathrm{IMP} = 0.7$.
During evaluation, we limit the maximum action sequence length to 25.

For all experiments, we keep 5\% of the training data as held-out from parameter updates and used as a validation set.
We use patience for stopping during pretraining of the plan predictor and the action generator (Section~\ref{sec:learn:pretrain}). 
We start with a patience of 10, which increases by a factor of 1.01 each time the stopping metric improves on the validation set.
For plan prediction training, we use patience on the validation set accuracy of predicted goal locations.
We compute goal location predictions by finding all positions $\position$ such that $p(\mathrm{GOAL} = 1 \mid \position, \state_1, \sentence) \geq 0.5$.
For action generation, we stop when card-state accuracy reaches a maximum on the validation set.
For fine-tuning (Section~\ref{sec:learn:finetune}), we stop training after $25$ epochs, and choose the epoch that maximizes the proportion of points scored computed using cascaded evaluation (Section~\ref{sec:cascaded}) on the validation set.

\paragraph{\system{Seq2seq+attn} Baseline}
We embed the sentence tokens into 64-dimensional vectors, and compute a sentence representation using a single-layer RNN with 64 LSTM hidden units.
We embed each position in the environment with a learned embedding function $\phi^\mathcal{S}$ mapping to a vector of size 32.
The resulting feature map is put through four convolutional layers separated by leaky ReLU non-linearities.
Each convolutional layer has a stride of two and divides the number of channels in half.
The output of the last convolutional layer is flattened to a vector.

We initialize the decoder hidden state to a zero-vector.
In each timestep we pass in the concatenation of the embedding of the previous output, the embedded environment vector, and the previous result of the attention computation on the sentence.
We take the initial attention result to be a zero vector.
We compute the attention over the sentence hidden states using the dot product of hidden state with the current hidden state in the decoder RNN.
The resulting attention state is concatenated with the decoder hidden state and the embedded environment vector, put through a leaky ReLU non-linearity, and and finally through a single fully-connected layer to predict probabilities over actions.

We train the model using teacher forcing and apply the same learning rate, optimizer, and stopping criteria as the fine-tuning experiments.

\end{document}